# Dissecting Motion-Prior Regularization for Data-Scarce Robotic Insertion

Ning Hu[1*], Shuai Li[2], and Jindong Tan[1]

[1] Department of Mechanical, Aerospace, and Biomedical Engineering, University of Tennessee, Knoxville, TN 37996, USA

[2] Engineering School of Sustainable Infrastructure and Environment, University of Florida, Gainesville, FL 32611, USA

[*] Corresponding author: Ning Hu (ning.hu.666@gmail.com)

**Abstract—**This study asks whether training-time motion-prior regularization can improve insertion success when a diffusion policy is learned from only 15 demonstrations. Minimum jerk discourages abrupt changes in predicted translational acceleration; speed-curvature regularization instead couples movement speed to path geometry. These are candidate mechanisms for task completion, not safety guarantees. We compare the priors individually and jointly, neither prior, and generic smoothness, with 80 real-robot trials per setting pooled over four recorded condition classes. Joint and minimum-jerk-only settings each achieved 70/80 successes (87.5%), versus 69/80 (86.3%) for speed-curvature only, 66/80 (82.5%) for neither prior, and 67/80 (83.8%) for generic smoothness. Success rates and Wilson 95% confidence intervals are visualized for direct comparison. Joint regularization exceeded neither by 5.0 percentage points but provided no observed gain over minimum jerk alone. The results motivate minimum jerk as the simpler candidate for replication, without establishing synergy, biomechanical specificity, improved safety, or distribution-shift robustness.

## I. Introduction

Precise insertion couples perceptual uncertainty, tight geometric tolerance, and contact dynamics. Diffusion Policy models multimodal robot behavior as conditional denoising over action sequences [1], while force-aware and slow-fast policies improve reactivity during contact [2], [3]. Data-efficient insertion has also benefited from object-centric pose representations [4] and geometric equivariance [5]. These advances leave a narrower training question: when only 15 demonstrations are available, do explicit motion-prior penalties improve successful insertion?

The primary objective is task success, measured by completion of the unplug-transfer-insert sequence. Motion regularity is a hypothesized means to that objective, not a measured endpoint in this study. Robustness would require condition-specific evaluation, and safety would require direct safety outcomes or verified constraints. Neither is inferred from the pooled success rate or from the mere use of a smoothness penalty.

Smooth action generation is not itself new. LiPo applies jerk-minimizing post-optimization to learned action chunks at inference time [6], and Frequency Guidance Operator (FGO) guides denoising through sub-frequency manifolds [7]. Our intervention applies minimum-jerk and speed-curvature penalties during policy training and examines their contributions through component and generic-smoothing controls. These established human-motion regularities [8]–[10] motivate candidate inductive biases; their biological origin does not establish their suitability for insertion.

We ask how the two priors, individually and jointly, affect pooled insertion success. The contributions are: (i) a training objective with distinct rationales for the two priors; (ii) a five-setting diagnostic with integer trial accounting and graphical confidence intervals; and (iii) a component-wise interpretation separating observed success differences from unverified mechanisms, synergy, robustness, and safety.

## II. Policy and Regularization

### A. Policy Interface

At time $t$, the policy observes

$$\mathcal{O}_t = \{I_t, T^B_{E,t}, \mathbf{w}_t, \mathcal{H}_{t-k:t}\},$$

where $I_t$ is an eye-in-hand RGB image, $T^B_{E,t}$ is the end-effector pose, $\mathbf{w}_t$ is the wrist force/torque measurement, and $\mathcal{H}$ is a short observation history. It predicts short-horizon actions

$$\mathbf{a}_t = [\Delta\mathbf{x}, \Delta\mathbf{R}_{6d}, g, c],$$

comprising Cartesian translation and rotation increments, gripper command $g$, and compliance code $c$. A ResNet-18 visual encoder conditions the diffusion head. Training uses 100k–200k iterations. Policy inference runs at 10–20 Hz, while interpolated Cartesian targets are executed at 50–200 Hz by compliant low-level control following standard impedance and hybrid force/position principles [11], [12]. Parameters not retained in the experimental record are not reported.

### B. Minimum-Jerk Prior: Rationale and Objective

Let $\mathbf{p}_0$ be the current Cartesian position and $\mathbf{p}_1, \dots, \mathbf{p}_H$ the predicted positions in an action chunk. The third finite difference is

$$D^{(3)}\mathbf{p}_k = \mathbf{p}_{k+3} - 3\mathbf{p}_{k+2} + 3\mathbf{p}_{k+1} - \mathbf{p}_k,$$

and the minimum-jerk penalty is

$$\mathcal{L}_J = \frac{1}{H-2}\sum_{k=0}^{H-3}\left\|D^{(3)}\mathbf{p}_k\right\|_2^2.$$

For uniformly spaced samples, dividing the third finite difference by the cube of the sampling interval approximates jerk, the third time derivative of position. The penalty therefore discourages abrupt changes in predicted translational acceleration within a chunk. Motivated by minimum-jerk movement models [8], [10], we hypothesize that this bias can suppress unnecessary short-scale action variation when demonstrations are scarce and make near-contact approach and alignment easier to execute. This is a proposed link to task completion, not an experimentally verified reduction in contact disturbance.

The penalty does not directly constrain rotation, inter-chunk discontinuities, or contact force. Excessive smoothing could also weaken a rapid correction needed for insertion. Its weight must therefore balance regularity against imitation fidelity; minimizing jerk alone does not guarantee task success or safe interaction.

### C. Speed-Curvature Prior: Rationale and Objective

For interior samples, the speed-curvature regularizer follows the two-thirds power-law form

$$\mathcal{L}_C = \frac{1}{H-1}\sum_{k=1}^{H-1}\left(\tilde{v}_k - \tilde{\alpha}\tilde{\kappa}_k^{-1/3}\right)^2,$$

where $\tilde{v}_k$ and $\tilde{\kappa}_k$ are normalized speed and curvature and $\tilde{\alpha}$ is the fitted scale.

The inverse one-third exponent expresses the two-thirds power law in terms of tangential speed and path curvature [9], [10]. Unlike minimum jerk, this term couples timing to geometry: it favors lower speed on more strongly curved path segments rather than directly penalizing changes in acceleration. The hypothesis is that such coordination may assist approach or reorientation. It need not help a contact-driven pause or nearly straight insertion, for which path curvature alone may not specify the appropriate speed.

TABLE I. FIVE-SETTING REGULARIZATION DESIGN

| Setting | Active regularization | Diagnostic role |
|---|---|---|
| Full | Minimum jerk + speed-curvature | Joint package |
| Jerk only | Minimum jerk | Jerk component |
| Speed-curvature only | Speed-curvature | Curvature component |
| Neither | Neither component | Factorial reference |
| Generic smoothness | Non-biomechanical smoothing | Specificity control |

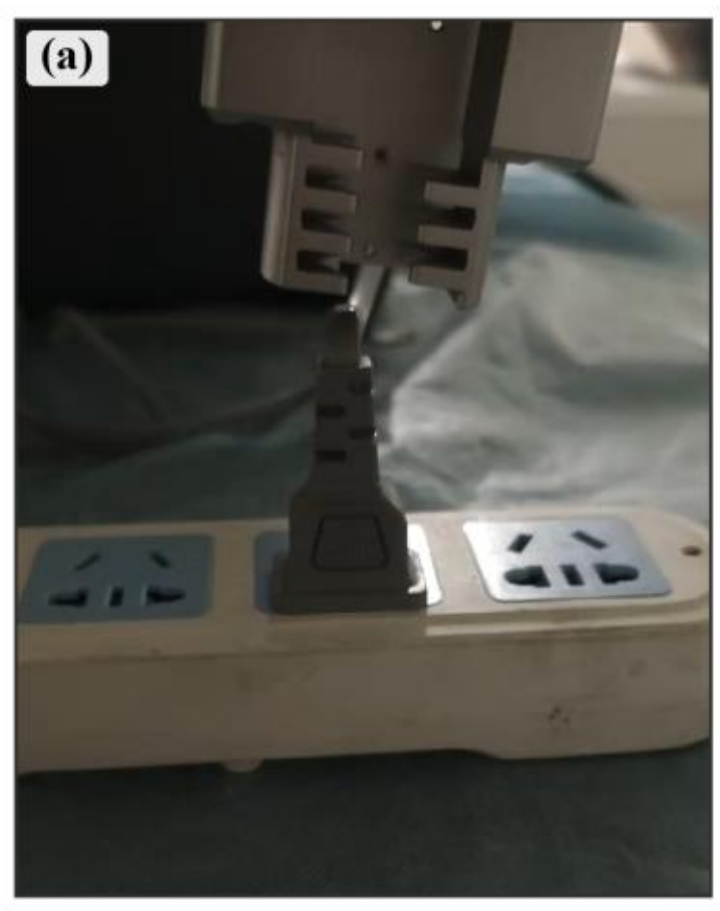

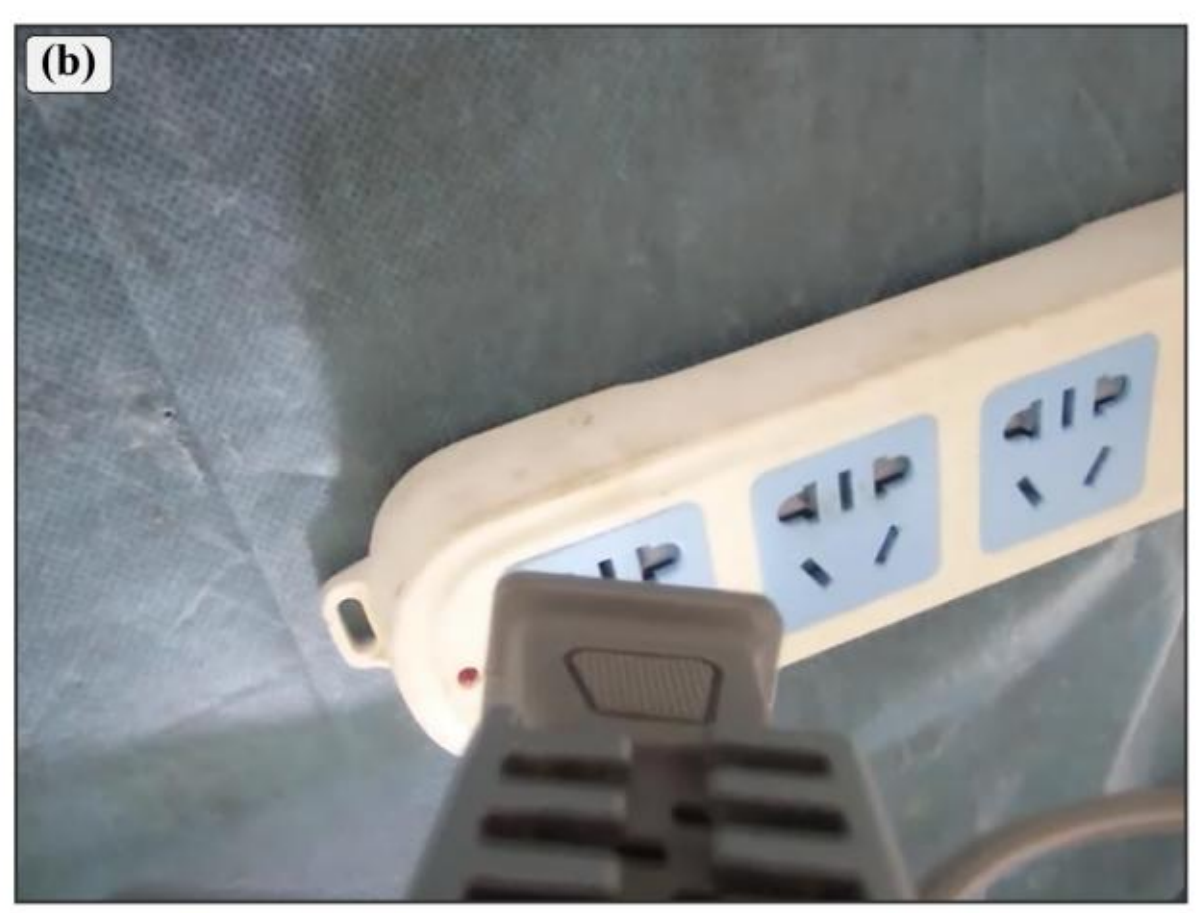

Fig. 1. Representative real-robot images from the archived experimental record: (a) frontal near-contact view of the gripper-held plug and socket; (b) oblique pre-insertion view showing the power-strip target. These images document task geometry, not separate task families or trial outcomes.

These different objectives need not yield additive gains. Their biases may overlap on some trajectories or compete with task-required corrections on others. The inverse-curvature form also needs a numerical convention near zero curvature. The retained manuscript does not specify that convention, the normalization details, or the fitted-scale procedure; no such implementation details are reconstructed from the aggregate outcomes.

*D. Joint Training and Matched Controls*

The complete training objective is

$$\mathcal{L} = \mathcal{L}_{\text{diff}} + \lambda_{\text{BC}}\mathcal{L}_{\text{BC}} + \lambda_J\mathcal{L}_J + \lambda_C\mathcal{L}_C + \lambda_{\text{eqv}}\mathcal{L}_{\text{eqv}}.$$

Here, the diffusion and auxiliary behavior-cloning losses retain the imitation objective, while the scalar weights control the additional penalties. The transformation-consistency term applies bounded SE(3) transformations consistently to observations and action increments. It is a training constraint rather than an architecturally equivariant network. Supporting perception, augmentation, compliance, and finite-state execution components are held fixed across the five settings. Exact unreported loss weights and implementation parameters remain unavailable.

Table I defines the matched regularization settings. The first four form the component comparison; generic smoothness probes whether any observed association is specific to the selected motion priors. The available record does not specify the generic penalty or establish matched regularization strength, so this is a limited specificity control. Fig. 1 documents the physical task; quantitative outcomes are reported separately in Fig. 2.

## III. Experimental Protocol and Results

*A. Platform and Trial Accounting*

Experiments use an AUBO-i5 six-degree-of-freedom manipulator, an eye-in-hand monocular RGB camera, a wrist-mounted six-axis force/torque sensor, and a parallel gripper. Training uses 15 teleoperated demonstrations of an unplug-transfer-insert sequence and no simulator pretraining. Evaluation spans Seen, Geometry, Pose, and Lighting condition classes, with exactly 20 trials per setting-condition and 80 pooled trials per setting. These are condition classes within one task family, not four independent tasks. Their perturbation magnitudes were not retained and are not treated as calibrated stress levels.

Success is the primary and only analyzed endpoint. A trial is successful when the full sequence completes within the task budget, the plug is seated and remains stable, and no safety abort occurs. Exact task-budget, geometric-tolerance, and dwell-time values are unavailable. An abort contributes to task failure; this composite definition does not separately quantify safety, contact force, or failure severity.

*B. Outcome Display and Statistical Scope*

Fig. 2 presents integer counts, derived success rates, and two-sided Wilson 95% confidence intervals in a common point-and-interval plot. The intervals are nominal binomial summaries of pooled outcomes, not estimates of training-seed variability or condition-specific performance. Condition-level success counts and the paired trial ledger are unavailable, so independence, pairing, and run-to-run variation cannot be verified. The analysis is consequently descriptive, with no paired tests or distribution-shift robustness estimates. Interval overlap is not used as a test of between-setting differences or equivalence.

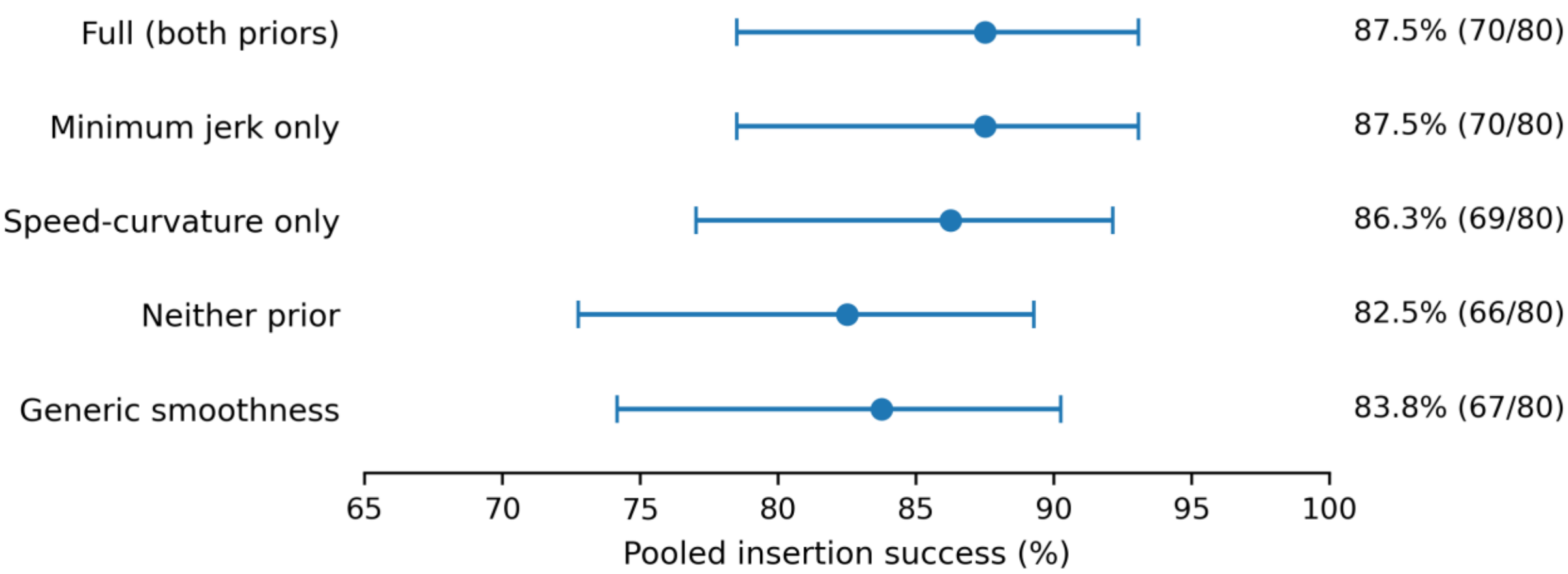

Fig. 2. Pooled insertion success across the five regularization settings. Points show success rates; horizontal error bars show two-sided Wilson 95% confidence intervals. Labels give the rate and successful/total trials. Each setting pools 20 trials from each of Seen, Geometry, Pose, and Lighting. The horizontal axis spans 65–100% for readability. Full and jerk only overlap exactly (70/80 each); these marginal intervals are not confidence intervals for pairwise differences.

### C. Component-Wise Success Results

The full and jerk-only settings each achieved 70/80 successes (87.5%), four more than neither prior (66/80; 82.5%; +5.0 percentage points). Speed-curvature only achieved 69/80 (86.3%), three more than neither (+3.75 points). Generic smoothness achieved 67/80 (83.8%); full and jerk only each exceeded it by three successes (+3.75 points). The Wilson intervals overlap substantially, and the small count differences are interpreted descriptively.

The conditional comparisons distinguish the priors. Adding minimum jerk to neither yielded +5.0 points, whereas adding it to speed-curvature yielded +1.25 points (70/80 versus 69/80). Adding speed-curvature to neither yielded +3.75 points, but adding it to minimum jerk yielded 0 points (70/80 versus 70/80). Thus, speed-curvature was numerically above the no-prior reference, yet offered no observed incremental benefit once minimum jerk was present. The tie is not evidence of equivalence.

For the 2 × 2 component comparison, averaging each component's success-rate difference over the other component's two levels gives descriptive main effects of +3.125 percentage points for minimum jerk and +1.875 points for speed-curvature. The difference-in-differences interaction is −3.75 points: the joint observed gain is smaller than the sum of the two individual gains. These are decompositions of aggregate proportions, not inferential factorial estimates or proof of antagonism.

For context only, an earlier archived system comparison recorded 70/80 for the full system, 67/80 for a platform-adapted FoAR baseline [2], 58/80 for Diffusion Policy [1], and 49/80 for vision-only behavior cloning. That comparison was not a concurrent randomized factorial run and is excluded from Fig. 2 and from component attribution.

## IV. Discussion

### A. What the Success Results Support

The results address task completion under the retained protocol, not general robustness or safety. Minimum jerk alone matched the highest observed success count without the additional speed-curvature term. This makes it a parsimonious candidate for replication rather than an established superior policy. The joint-versus-neither comparison is compatible with a useful regularization signal, but the component comparison provides no observed added value for joint regularization. The negative interaction estimate does not support a synergy claim.

The generic-smoothness comparison also matters: the three-success difference is insufficient to establish that human-movement structure, rather than regularization more generally, explains performance. A mechanistic claim would require trajectory measurements and a fully specified, strength-matched smoothing control, not just a full model and a joint-removal ablation.

### B. Why the Two Priors May Affect Success Differently

Minimum jerk penalizes temporal variation in predicted acceleration without prescribing a specific speed for each local path shape. Speed-curvature regularization instead assumes that local geometry should organize speed. As an interpretation, the former may be more broadly useful when insertion requires smooth approach interspersed with contact-driven adjustments, whereas the latter may add little if the useful trajectories are mostly straight or their timing is dictated by contact rather than curvature.

Other explanations remain possible: the priors may impose partly redundant biases, the curvature term may be sensitive to near-zero-curvature handling, or the chosen weights may trade imitation fidelity against regularity differently. These are hypotheses, not measured failure mechanisms. No retained jerk, curvature, force, trajectory-quality, or contact-stability outcomes distinguish them. In particular, the results do not show that speed-curvature regularization is universally ineffective, or that minimum jerk improved success by demonstrably reducing jerk.

### C. Limitations and Targeted Replication

This is an archival diagnostic pilot covering one robot, one task family, 15 demonstrations, and binary success. Missing trial order, checkpoint identifiers, per-condition counts, and exact unreported hyperparameters limit reproducibility and prevent verified independence, causal attribution, or shift-specific conclusions. No improvement in execution time, contact stability, or safety is established. The evaluation at one demonstration count also does not establish a data-efficiency scaling advantage.

A targeted replication should retain trial-level condition and outcome records, fixed success and abort criteria, exact regularizer definitions and weights, and independent training runs. Measuring executed jerk alongside success would test the proposed minimum-jerk mechanism; retaining path curvature and contact-phase outcomes would test when speed-curvature helps or conflicts with the task. Condition-resolved and safety-specific outcomes would be needed before expanding the claims beyond pooled task completion.

## V. Conclusion

With 15 demonstrations and 80 pooled real-robot trials per setting, full and minimum-jerk-only regularization each achieved 70 successes, compared with 69 for speed-curvature only, 66 for neither prior, and 67 for generic smoothness. Minimum jerk alone matched the best observed task success, while speed-curvature provided no observed incremental gain when added to it. The point-and-interval comparison supports a modest descriptive success signal, not demonstrated synergy, biomechanical specificity, robustness, or safety. Minimum jerk is therefore the simpler candidate for a preregistered, trial-level replication.

**Data and code statement:** The aggregate counts used here are reported in the manuscript. The original trial-level ledger and exact values of unreported implementation parameters are unavailable.

**Funding and conflict of interest:** This research received no specific funding. The authors declare no conflicts of interest relevant to this work.

**AI assistance disclosure:** AI-assisted tools were used for language editing, consistency checks, visualization of reported aggregate results, and document formatting; the authors remain responsible for the scientific content and reported results.

## References

[1] C. Chi, Z. Xu, S. Feng, E. Cousineau, Y. Du, B. Burchfiel, R. Tedrake, and S. Song, “Diffusion Policy: Visuomotor Policy Learning via Action Diffusion,” Robotics: Science and Systems, 2023.

[2] Z. He, H. Fang, J. Chen, H.-S. Fang, and C. Lu, “FoAR: Force-Aware Reactive Policy for Contact-Rich Robotic Manipulation,” IEEE Robotics and Automation Letters, 2025.

[3] H. Xue, J. Ren, W. Chen, G. Zhang, Y. Fang, G. Gu, H. Xu, and C. Lu, “Reactive Diffusion Policy: Slow-Fast Visual-Tactile Policy Learning for Contact-Rich Manipulation,” Robotics: Science and Systems, 2025.

[4] H. Sun, S. Liu, Y. Wang, Z. Zhou, S. Wang, H. Yang, J. Sun, and Q. Cao, “Exploring Pose-Guided Imitation Learning for Robotic Precise Insertion,” arXiv:2505.09424, 2025.

[5] J. Yang, Z.-A. Cao, C. Deng, R. Antonova, S. Song, and J. Bohg, “EquiBot: SIM(3)-Equivariant Diffusion Policy for Generalizable and Data Efficient Learning,” Conference on Robot Learning, 2024.

[6] D. Son and S. Park, “LiPo: A Lightweight Post-optimization Framework for Smoothing Action Chunks Generated by Learned Policies,” arXiv:2506.05165, 2025.

[7] J. Wang, “Frequency-Guided Action Diffusion via Sub-Frequency Manifold Traversal,” arXiv:2605.27919, 2026.

[8] T. Flash and N. Hogan, “The Coordination of Arm Movements: An Experimentally Confirmed Mathematical Model,” Journal of Neuroscience, vol. 5, no. 7, pp. 1688–1703, 1985.

[9] F. Lacquaniti, C. Terzuolo, and P. Viviani, “The Law Relating the Kinematic and Figural Aspects of Drawing Movements,” Acta Psychologica, vol. 54, nos. 1–3, pp. 115–130, 1983.

[10] P. Viviani and T. Flash, “Minimum-Jerk, Two-Thirds Power Law, and Isochrony: Converging Approaches to Movement Planning,” Journal of Experimental Psychology: Human Perception and Performance, vol. 21, no. 1, pp. 32–53, 1995.

[11] N. Hogan, “Impedance Control: An Approach to Manipulation: Part II—Implementation,” Journal of Dynamic Systems, Measurement, and Control, vol. 107, no. 1, pp. 8–16, 1985.

[12] M. H. Raibert and J. J. Craig, “Hybrid Position/Force Control of Manipulators,” Journal of Dynamic Systems, Measurement, and Control, vol. 103, no. 2, pp. 126–133, 1981.